\documentclass[sigconf]{acmart}
\usepackage{xspace}
\usepackage{graphicx}
\usepackage{booktabs}
\usepackage{tabularx}
\usepackage{array}
\usepackage{amsmath}
\usepackage{xurl}
\newcommand{\system}{\textsc{RecGroupMemory}\xspace}
\acmConference[WWW '27]{ACM Web Conference 2027}{May 10--14, 2027}{Dublin, Ireland}
\acmBooktitle{Review manuscript for the ACM Web Conference 2027}
\acmYear{2027}
\copyrightyear{}
\setcopyright{none}
\acmDOI{}
\acmISBN{}
\usepackage{enumitem}

\title[RecGroupMemory]{RecGroupMemory: Feedback-Aware Cohort Memory for E-Commerce Recommendation}
\author{Wendong Zhang}
\authornote{Corresponding author.}
\affiliation{\institution{}\city{}\country{}}
\email{xianzhi.zwd@alibaba-inc.com}

\author{Weipeng Huang}
\affiliation{\institution{}\city{}\country{}}
\email{weipeng.hwp@taobao.com}

\author{Zhixin Ma}
\affiliation{\institution{}\city{}\country{}}
\email{mazhixin.mzx@taobao.com}

\author{Bin Zhang}
\affiliation{\institution{}\city{}\country{}}
\email{tianji.zb@taobao.com}

\author{Dimin Wang}
\affiliation{\institution{}\city{}\country{}}
\email{dimin.wdm@taobao.com}

\begin{document}
\begin{abstract}
Cohort memory can support LLM-based recommendation planning through group-level user profiles and advice. As outcome feedback becomes available, this guidance may need to be updated while preserving the underlying observations. We present \system, a workflow that provides this memory to an orchestration LLM and separates memory construction from feedback-based revision. It uses feedback from the previous day to revise decision guidance, while validation checks keep selected source values unchanged. Online comparisons were conducted among Taobao users with low purchasing power selected by the business team. Relative to a no-memory baseline, memory without feedback and memory with feedback showed increases of $1.58\%$ and $1.78\%$, respectively, in the number of guided transaction users.
\end{abstract}

\maketitle

\section{Introduction}
Cohort memory summarizes observations about users with shared characteristics to support an orchestration LLM in planning Web recommendation workflows across requests. Delayed browsing and transaction outcomes may challenge earlier guidance without changing the cohort definition, observation window, or recorded metrics. Updating this memory requires revising recommendation judgments while preserving their observational basis. Unconstrained regeneration risks changing the recorded observations. The same observations can support different planning judgments as later outcomes arrive. Keeping observations separate from guidance makes these revisions explicit: an update changes the interpretation of a cohort record while retaining the evidence against which that interpretation can be checked.

Prior recommendation work models user interests and behavioral histories~\cite{zhou2018din,kang2018sasrec,pi2019mimn,chen2018usermemory}, uses language models for recommendation~\cite{geng2022p5,bao2023tallrec}, and updates profiles or retrieves user-history memory~\cite{bang2025pure,wang2025automr}. Agent memory research studies reflection and memory management~\cite{park2023generative,packer2023memgpt,chhikara2025mem0}, as well as feedback-based revision~\cite{shinn2023reflexion,madaan2023selfrefine}. We focus on using external recommendation outcomes to revise cohort-level planning guidance.

Recommendation feedback depends on the policy that generated it~\cite{swaminathan2015crm,schnabel2016treatments}. Our workflow provides planning advice rather than learning a ranking policy, and linking feedback to memory through Memory ID does not establish causal attribution.

\begin{figure*}[t]
  \centering
  \includegraphics[width=0.90\textwidth]{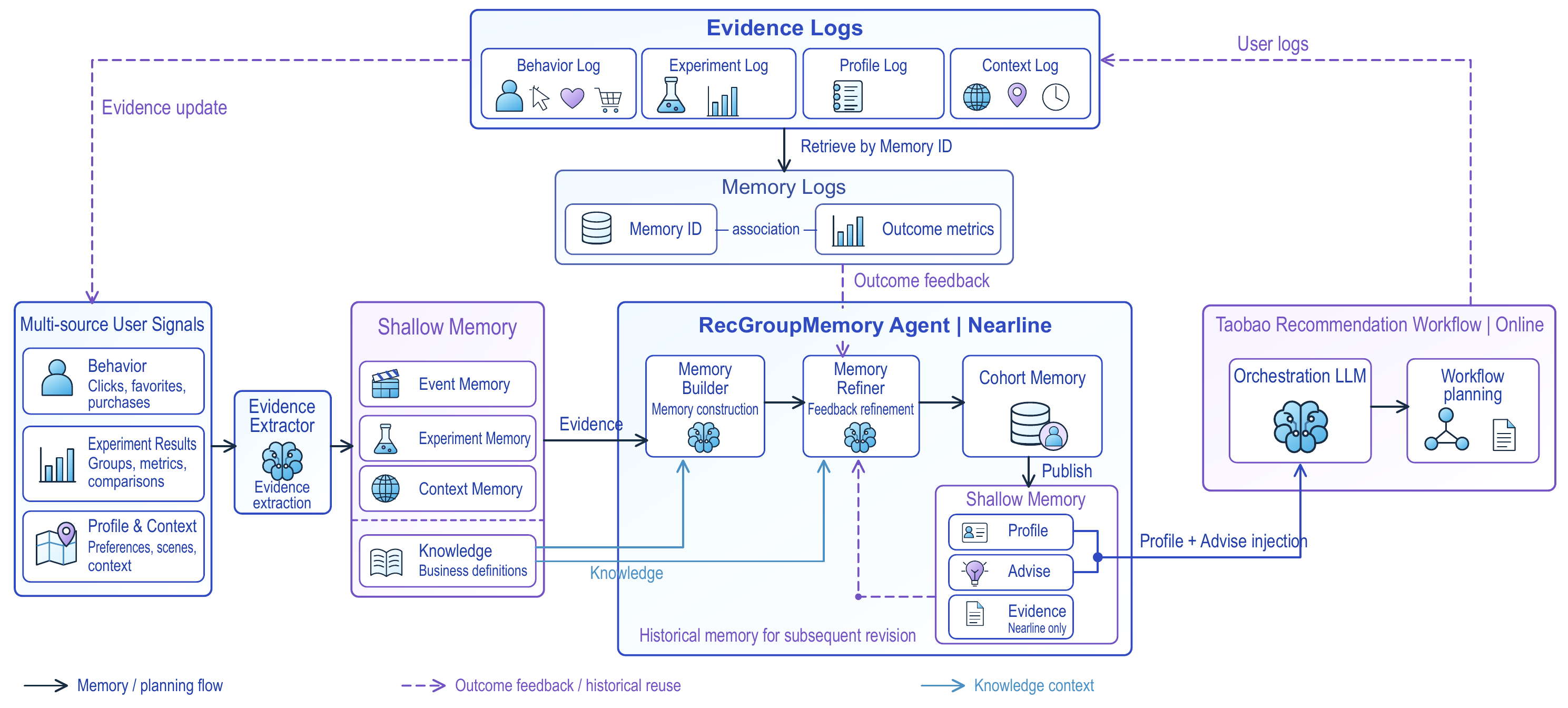}
  \caption{RecGroupMemory system overview. Multi-source evidence and online outcomes support nearline synthesis and feedback refinement. Published Profile and Advise guide the orchestration LLM; stored memory supports subsequent updates.}
  \Description{The Evidence Extractor organizes multi-source observations into event, experiment, and context memory in Shallow Memory. The Memory Builder constructs cohort memory and the Memory Refiner revises its guidance; knowledge supports both. Evidence Logs link to Memory Logs by Memory ID and provide outcomes to Memory Refiner. Refined Cohort Memory is published to Shallow Memory, which retains Profile, Advise, and Evidence. Only Profile and Advise feed the orchestration LLM for workflow planning. Online logs return to Evidence Logs, which also update user signals. Stored memory feeds Memory Refiner for feedback refinement in the next cycle.}
  \label{fig:overview}
\end{figure*}

We propose \system (Figure~\ref{fig:overview}), with the following contributions:
\begin{itemize}[labelindent=0pt,leftmargin=*,labelsep=0.5em]
\item The framework separates memory construction from feedback revision and provides an orchestration LLM with cohort profiles (Profile) and advice (Advise) derived from multiple observation sources.
\item The revision procedure uses outcome feedback linked by Memory ID to update decision content while preserving selected source values. Eligibility checks, validation, and fallback to the original record control the update.
\item Online comparisons among business-selected Taobao users with low purchasing power show increases in guided transaction users of $1.58\%$ for memory without feedback and $1.78\%$ for memory with feedback, each relative to its corresponding no-memory baseline.
\end{itemize}

\section{The RecGroupMemory Framework}
\label{sec:method}

The framework separates memory construction, feedback revision, and online planning. The Evidence Extractor organizes observations from multiple sources in Shallow Memory. The Memory Builder combines these observations into a cohort memory record, and the Memory Refiner reassesses its guidance when linked outcome feedback becomes available. The online orchestration LLM uses the published Profile and Advise to plan the recommendation workflow. Feedback revision updates the planning context without scoring items or executing downstream recommendation stages.

Each update operates on a cohort memory record rather than an individual interaction or a model parameter. The record contains observations for a defined user group and time window, along with guidance derived from them. This guidance can be revised while selected source values remain unchanged.

\subsection{Memory Inputs and Feedback}

The Evidence Extractor extracts observations from multiple sources to form event, experiment, and context memory in Shallow Memory. The Memory Builder combines these observations into cohort memory, and the Memory Refiner revises its guidance using outcome feedback (Figure~\ref{fig:overview}).

The feedback node reads the current response $C_T$ from date partition $T$ and left-joins previous-day feedback $F_{T-1}$ by Memory ID. Revision requires exactly one matching feedback row and upstream flags confirming profile validity, baseline matching, and feedback eligibility. Records with no unique match, failed eligibility checks, or invalid source responses retain $C_T$. Eligibility does not guarantee sufficient evidence for every objective. The prompt considers missing metrics when selecting a direction and recording uncertainty.

For Memory ID $i$ and metric $q$, upstream feedback is $\delta_{i,q}=q_i/q_{\mathrm{base}(i)}-1$, where $q_{\mathrm{base}(i)}$ is computed over all traffic with the same profile in the platform's first-guess recommendation feed. This observational feedback covers activity, browsing, item-detail views, and transactions. The prompt uses this signed relative change directly: zero indicates no observed change, whereas a missing value indicates absent evidence and must not be treated as neutral or positive feedback.

For an eligible record, a historical profile $H_T$ retrieved from Shallow Memory provides optional context. Retrieval checks the cohort and any available version metadata; $H_T$ need not come from the immediately preceding partition. Revision can proceed without history if retrieval fails. Budgeted knowledge retrieval supplies terminology and business context. The prompt uses history to compare earlier advice and changes over time, but does not allow historical measurements to replace current observations or override new feedback. $C_T$ remains the revision baseline.

These inputs have distinct roles: $C_T$ identifies what is being revised, $F_{T-1}$ supplies outcome evidence, and $H_T$ supplies optional historical context. Available history cannot compensate for missing or ambiguous feedback matching. This separation prevents the availability of retrieval from being confused with eligibility for revision.

\subsection{Constrained Decision Revision}
\begin{figure}[t]
  \centering
  \includegraphics[width=\columnwidth]{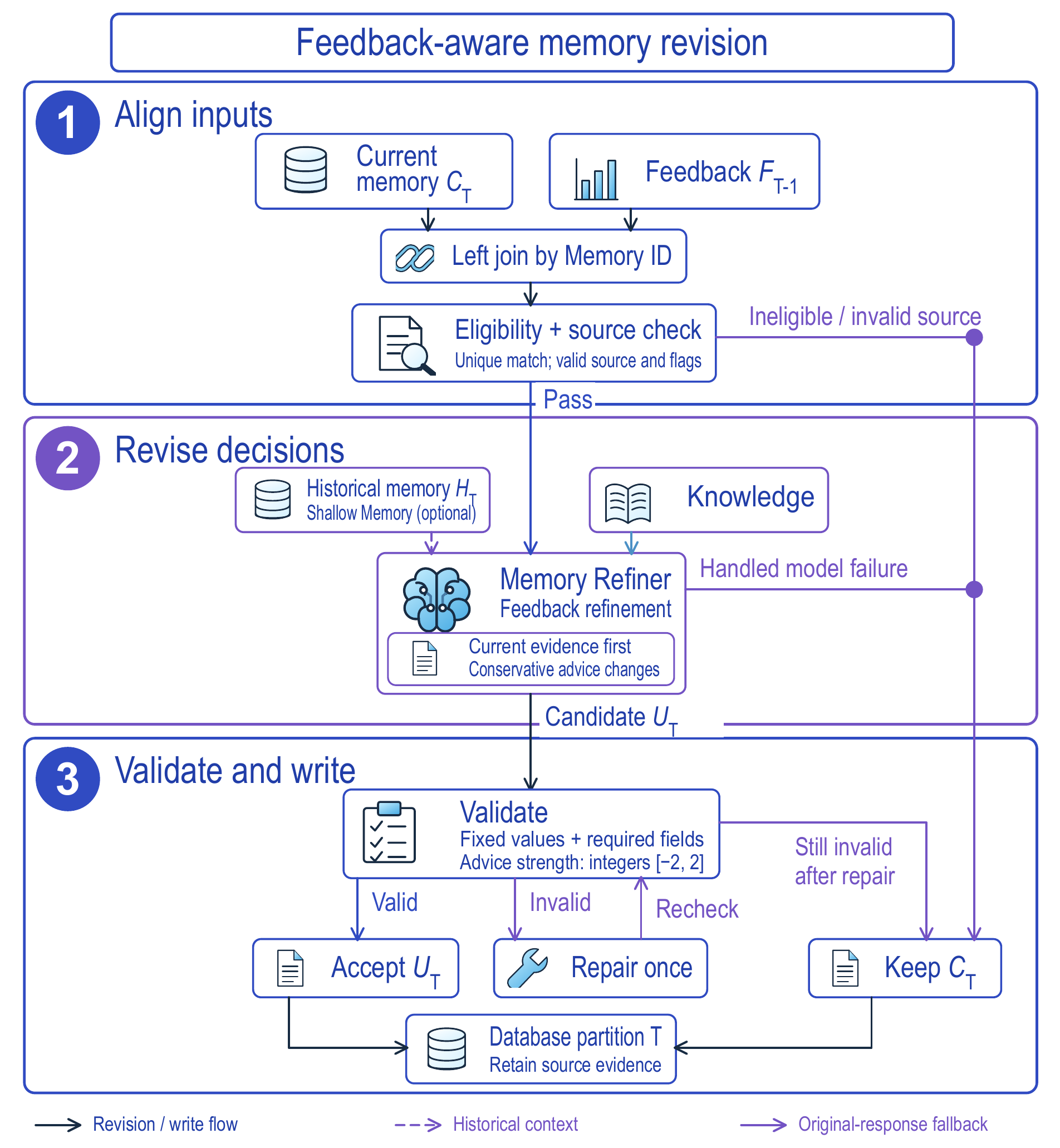}
  \caption{Feedback-aware revision with at most one repair and original-response fallback. Fixed values are checked against $C_T$. Direction-to-strength checks depend on configuration; semantic policies remain prompt-based.}
  \Description{Three outlined stages on a white background show input alignment, decision revision, and validation with writing. Current memory C_T and previous-day feedback F_(T-1) are left-joined by Memory ID and checked for eligibility and source validity. Passing records enter Memory Refiner, supported by optional historical memory H_T from Shallow Memory, knowledge, and prompt policies. Candidate U_T enters validation; an invalid candidate permits one repair and recheck. A valid candidate is accepted. Ineligible or invalid-source records, still-invalid candidates, and handled model failures retain C_T. Accepted and retained responses are written to database partition T with source evidence preserved. Direction-to-strength checks depend on configuration; semantic policies remain prompt-based.}
  \label{fig:contract}
\end{figure}

Figure~\ref{fig:contract} shows how the Memory Refiner revises decision content while validation preserves selected source values. Seven values must match $C_T$: schema version, cohort, recommender version, memory scope, segment definition, observation window, and aggregated metrics. Profile and Advise must remain nonempty strings. The prompt permits changes to their content, structured advice, profile confidence, uncertainties, and change metadata. The validator checks selected values and required structure, rather than enforcing a complete whitelist of editable JSON fields. The fixed values identify the cohort, observation window, and measured behavior to which the advice applies. Preserving them prevents an apparent improvement in the narrative from being obtained by changing the population or measurements. Editable decision content instead describes how those observations should inform subsequent planning.

Profile summarizes the cohort's current funnel performance and what the feedback suggests about the existing strategy. Advise specifies adjustments for particular objectives, their rationale, and rollback conditions. The prompt records missing metrics and data-quality concerns in uncertainties and summarizes changes from $C_T$ in change metadata. Confidence and uncertainty qualify the interpretation. The conversion-prioritized prompt requests a concise profile and ordered advice, with final discrete strengths stored in structured advice.

Advice strengths are integers in $\{-2,-1,0,1,2\}$ for four objectives: click-through rate (CTR), item-page views (IPV), conversion rate (CVR), and gross merchandise value (GMV). The intended update is
\begin{equation}
 a'_q=\operatorname{clip}_{[-2,2]}(a_q+\Delta_q),
 \qquad \Delta_q\in\{-1,0,1\},
 \label{eq:strength}
\end{equation}
where $a_q$ is taken from $C_T$, or set to zero for legacy records without structured advice. The action \emph{revert to previous} keeps the strength in $C_T$ unchanged rather than restoring $H_T$.

The candidate stores the final strength $a'_q$, not the increment $\Delta_q$. These values express advice strength, not calibrated probabilities or ranking-score multipliers. Clipping can leave the stored strength unchanged at a bound, so the rationale helps distinguish the requested adjustment from its numerical result.

Objective priorities are configurable. The conversion-prioritized prompt emphasizes CVR and suppresses GMV adjustment; the browsing-prioritized prompt emphasizes IPV with transaction safeguards. Validation checks advice structure, integer type, and range; only the browsing configuration also checks direction-to-strength consistency. Conservative choices and most semantic constraints depend on the prompt. Equation~\ref{eq:strength} therefore describes the intended update, not a guarantee enforced across all configurations.

The conversion-prioritized prompt permits CVR strengthening only when the relative changes in both transaction UV and order count are available and strictly positive. Positive intermediate browsing feedback cannot substitute for either outcome. If either relative change is missing or non-positive, the prompt calls for conservative conversion advice. It separates assessment of the previous strategy from selection of the next adjustment: retaining the current advice under insufficient evidence does not imply that the strategy succeeded.

For example, positive browsing feedback accompanied by missing transaction UV cannot satisfy the conversion-strengthening condition. The record can remain eligible for revision while that particular adjustment lacks support. The resulting Profile can describe the observed browsing response, and Advise can retain a conservative conversion position while recording the missing outcome in uncertainties.

\subsection{Validation and Workflow Integration}
Let $U_T$ denote the generated candidate after at most one repair attempt, $O_T$ the output response, and $e_T=1$ indicate that the record is eligible and its source response is valid. For handled record-level outcomes,
\begin{equation}
 O_T=\begin{cases}
 U_T, & e_T=1\ \land\ \mathcal{V}(U_T,C_T)=1,\\
 C_T, & \text{otherwise},
 \end{cases}
 \label{eq:update}
\end{equation}
where $\mathcal{V}$ checks the fixed source values and required output fields. Eligibility permits revision; acceptance requires the candidate to pass validation.

The Memory Refiner returns a complete JSON response rather than a patch. The feedback node parses this response and checks fixed-value equality, required text fields, and structured advice. An invalid candidate may undergo one repair attempt, which asks the model to restore valid JSON and fixed values while retaining its conclusions and advice text. The repaired candidate is validated again against $C_T$, but these checks do not establish semantic equivalence. Invalid source records bypass generation, and handled model or validation failures retain the original response.

An accepted update replaces the response and model metadata while retaining source evidence columns. The feedback node writes records to database partition $T$; publication to Shallow Memory occurs in a separate stage. The store retains Profile, Advise, and Evidence, but only Profile and Advise enter the orchestration LLM for workflow planning. Stored memory and online evidence support later updates. The revision node does not produce ranking weights. The database stores the revision output, whereas Shallow Memory makes the published record available to later planning and revision. A successful database write and a published memory record are therefore separate workflow outcomes. Figure~\ref{fig:overview} shows the publication boundary, while Figure~\ref{fig:contract} ends at the database write.

Before processing, source IDs must be nonempty and unique. After writing the records, the job checks that the output and input have the same record count and ID set. Per-record statuses distinguish accepted generation from ineligible input, invalid source, and handled model failures. Acceptance indicates successful validation, not necessarily changed advice or improved outcomes. Fallback preserves record coverage without establishing that the original advice is optimal. Publication failures remain the responsibility of the surrounding pipeline.

\section{Online Experiments}
\label{sec:experiments}

\subsection{Experimental Setup}

Online experiments were conducted among Taobao users with low purchasing power selected by the business team. The no-memory baseline $B$ uses an orchestration LLM to plan the recommendation workflow from request context. Existing components handle retrieval, ranking, and re-ranking.

Experiment E1 compares $B$ with memory $M$, which adds Profile and Advise generated by the Memory Builder to the planning context, without feedback revision. Experiment E2 compares its no-memory baseline with feedback-aware memory $F$, which uses the same memory interface and enables the Memory Refiner to revise these fields using previous-day outcomes. Each experiment reports relative changes against its own no-memory control.

\begin{table}[t]
\centering
\caption{Relative changes against each experiment's no-memory baseline.}
\label{tab:results}
\small
\begin{tabularx}{\columnwidth}{@{}Xrr@{}}
\toprule
Metric & $M/B$ (E1) & $F/B$ (E2) \\
\midrule
\textbf{Transaction UV} & $+1.58\%$ & $+1.78\%$ \\
Orders & $+0.80\%$ & $+2.05\%$ \\
PCVR & $+1.60\%$ & $+1.95\%$ \\
\midrule
Exposure UV & $+0.14\%$ & $+0.02\%$ \\
Click UV & $-0.42\%$ & $+0.04\%$ \\
IPV UV & $-0.50\%$ & $+0.17\%$ \\
\bottomrule
\end{tabularx}
\end{table}

\subsection{Results}

Table~\ref{tab:results} shows positive changes in transaction UV, orders, and PCVR for both configurations. Relative to their respective baselines, $M$ and $F$ increase transaction UV by $1.58\%$ and $1.78\%$, orders by $0.80\%$ and $2.05\%$, and PCVR by $1.60\%$ and $1.95\%$, respectively.

These transaction gains coincide with small changes in exposure UV: $+0.14\%$ for $M$ and $+0.02\%$ for $F$. For $M$, click UV and IPV UV decline by $0.42\%$ and $0.50\%$, despite the increases in transaction metrics. For $F$, these browsing metrics rise slightly, by $0.04\%$ and $0.17\%$. The observed improvements are therefore concentrated in transaction outcomes, while exposure and browsing volumes remain close to baseline.

Although $F$ reports larger transaction-related uplifts, the two configurations are compared with their respective baselines. These results do not directly estimate the incremental benefit of adding feedback to $M$.

\section{Discussion and Limitations}

Validation covers selected source values and output structure; semantic correctness still depends on the LLM and prompt instructions. Historical retrieval is not tied to a specific date or content hash, so Memory ID alone cannot identify the version previously used online. Fallback operates on individual records, and a failed partition write may still leave partial output.

The experiments evaluate complete workflow configurations on a selected user population. Deployment artifacts corresponding to the experiments are unavailable, limiting attribution to individual revision safeguards. Broader deployment and evaluation of these safeguards require feedback ablations under matched conditions, traces linking served memory to revisions, and runtime measurements. Natural-language memory and small cohorts also require access controls, retention limits, and aggregation safeguards. Evaluation should distinguish record eligibility, validation acceptance, changes in advice, and downstream outcomes. These events answer different questions about coverage, structural reliability, revision behavior, and recommendation performance. Recording them separately would make failures and unchanged advice easier to interpret without treating every accepted response as an effective update.

\balance
\section{Conclusion}

The framework integrates cohort memory construction, feedback revision, and LLM-based recommendation planning. It revises Profile and Advise while preserving selected source values, using validation and fallback to the original record to control updates.
\bibliographystyle{ACM-Reference-Format}
\bibliography{references}

@inproceedings{zhou2018din,
  title     = {Deep Interest Network for Click-Through Rate Prediction},
  author    = {Zhou, Guorui and Zhu, Xiaoqiang and Song, Chenru and Fan, Ying and Zhu, Han and Ma, Xiao and Yan, Yanghui and Jin, Junqi and Li, Han and Gai, Kun},
  booktitle = {Proceedings of the 24th ACM SIGKDD International Conference on Knowledge Discovery and Data Mining},
  pages     = {1059--1068},
  year      = {2018},
  doi       = {10.1145/3219819.3219823}
}

@inproceedings{pi2019mimn,
  title     = {Practice on Long Sequential User Behavior Modeling for Click-Through Rate Prediction},
  author    = {Pi, Qi and Bian, Weijie and Zhou, Guorui and Zhu, Xiaoqiang and Gai, Kun},
  booktitle = {Proceedings of the 25th ACM SIGKDD International Conference on Knowledge Discovery and Data Mining},
  pages     = {2671--2679},
  year      = {2019},
  doi       = {10.1145/3292500.3330666}
}

@inproceedings{kang2018sasrec,
  title     = {Self-Attentive Sequential Recommendation},
  author    = {Kang, Wang-Cheng and McAuley, Julian},
  booktitle = {2018 IEEE International Conference on Data Mining},
  pages     = {197--206},
  year      = {2018},
  doi       = {10.1109/ICDM.2018.00035}
}

@inproceedings{chen2018usermemory,
  title     = {Sequential Recommendation with User Memory Networks},
  author    = {Chen, Xu and Xu, Hongteng and Zhang, Yongfeng and Tang, Jiaxi and Cao, Yixin and Qin, Zheng and Zha, Hongyuan},
  booktitle = {Proceedings of the Eleventh ACM International Conference on Web Search and Data Mining},
  pages     = {108--116},
  year      = {2018},
  doi       = {10.1145/3159652.3159668}
}

@inproceedings{geng2022p5,
  title     = {Recommendation as Language Processing ({RLP}): A Unified Pretrain, Personalized Prompt \& Predict Paradigm ({P5})},
  author    = {Geng, Shijie and Liu, Shuchang and Fu, Zuohui and Ge, Yingqiang and Zhang, Yongfeng},
  booktitle = {Proceedings of the 16th ACM Conference on Recommender Systems},
  pages     = {299--315},
  year      = {2022},
  doi       = {10.1145/3523227.3546767}
}

@inproceedings{bao2023tallrec,
  title   = {{TALLRec}: An Effective and Efficient Tuning Framework to Align Large Language Model with Recommendation},
  author  = {Bao, Keqin and Zhang, Jizhi and Zhang, Yang and Wang, Wenjie and Feng, Fuli and He, Xiangnan},
  booktitle = {Proceedings of the 17th ACM Conference on Recommender Systems},
  pages   = {1007--1014},
  year    = {2023},
  doi     = {10.1145/3604915.3608857}
}

@misc{bang2025pure,
  title   = {{LLM}-based User Profile Management for Recommender System},
  author  = {Bang, Seunghwan and Song, Hwanjun},
  year    = {2025},
  eprint  = {2502.14541v2},
  eprinttype = {arXiv},
  archivePrefix = {arXiv},
  primaryClass = {cs.CL}
}

@inproceedings{park2023generative,
  title   = {Generative Agents: Interactive Simulacra of Human Behavior},
  author  = {Park, Joon Sung and O'Brien, Joseph and Cai, Carrie Jun and Morris, Meredith Ringel and Liang, Percy and Bernstein, Michael S.},
  booktitle = {Proceedings of the 36th Annual ACM Symposium on User Interface Software and Technology},
  pages   = {1--22},
  year    = {2023},
  doi     = {10.1145/3586183.3606763}
}

@misc{packer2023memgpt,
  title   = {{MemGPT}: Towards {LLM}s as Operating Systems},
  author  = {Packer, Charles and Wooders, Sarah and Lin, Kevin and Fang, Vivian and Patil, Shishir G. and Stoica, Ion and Gonzalez, Joseph E.},
  year    = {2023},
  eprint  = {2310.08560},
  eprinttype = {arXiv},
  archivePrefix = {arXiv},
  primaryClass = {cs.AI}
}

@inproceedings{shinn2023reflexion,
  title     = {Reflexion: Language Agents with Verbal Reinforcement Learning},
  author    = {Shinn, Noah and Cassano, Federico and Gopinath, Ashwin and Narasimhan, Karthik and Yao, Shunyu},
  booktitle = {Advances in Neural Information Processing Systems},
  volume    = {36},
  pages     = {8634--8652},
  year      = {2023},
  publisher = {Curran Associates, Inc.},
  doi       = {10.52202/075280-0377}
}

@inproceedings{madaan2023selfrefine,
  title     = {Self-Refine: Iterative Refinement with Self-Feedback},
  author    = {Madaan, Aman and Tandon, Niket and Gupta, Prakhar and Hallinan, Skyler and Gao, Luyu and Wiegreffe, Sarah and Alon, Uri and Dziri, Nouha and Prabhumoye, Shrimai and Yang, Yiming and Gupta, Shashank and Majumder, Bodhisattwa Prasad and Hermann, Katherine and Welleck, Sean and Yazdanbakhsh, Amir and Clark, Peter},
  booktitle = {Advances in Neural Information Processing Systems},
  volume    = {36},
  pages     = {46534--46594},
  year      = {2023},
  publisher = {Curran Associates, Inc.},
  doi       = {10.52202/075280-2019}
}

@misc{chhikara2025mem0,
  title   = {{Mem0}: Building Production-Ready {AI} Agents with Scalable Long-Term Memory},
  author  = {Chhikara, Prateek and Khant, Dev and Aryan, Saket and Singh, Taranjeet and Yadav, Deshraj},
  year    = {2025},
  eprint  = {2504.19413},
  eprinttype = {arXiv},
  archivePrefix = {arXiv},
  primaryClass = {cs.CL}
}

@inproceedings{swaminathan2015crm,
  title     = {Counterfactual Risk Minimization: Learning from Logged Bandit Feedback},
  author    = {Swaminathan, Adith and Joachims, Thorsten},
  booktitle = {Proceedings of the 32nd International Conference on Machine Learning},
  series    = {Proceedings of Machine Learning Research},
  volume    = {37},
  pages     = {814--823},
  year      = {2015},
  publisher = {PMLR},
  address   = {Lille, France},
  url       = {https://proceedings.mlr.press/v37/swaminathan15.html}
}

@inproceedings{schnabel2016treatments,
  title     = {Recommendations as Treatments: Debiasing Learning and Evaluation},
  author    = {Schnabel, Tobias and Swaminathan, Adith and Singh, Ashudeep and Chandak, Navin and Joachims, Thorsten},
  booktitle = {Proceedings of the 33rd International Conference on Machine Learning},
  series    = {Proceedings of Machine Learning Research},
  volume    = {48},
  pages     = {1670--1679},
  year      = {2016},
  publisher = {PMLR},
  address   = {New York, New York, USA},
  url       = {https://proceedings.mlr.press/v48/schnabel16.html}
}

@inproceedings{wang2025automr,
  title     = {Leveraging Memory Retrieval to Enhance {LLM}-based Generative Recommendation},
  author    = {Wang, Chengbing and Zhang, Yang and Zhu, Fengbin and Zhang, Jizhi and Shi, Tianhao and Feng, Fuli},
  booktitle = {Companion Proceedings of the ACM on Web Conference 2025},
  pages     = {1346--1350},
  year      = {2025},
  doi       = {10.1145/3701716.3715596}
}
\end{document}